\documentclass[11pt,a4paper]{article}
\usepackage[hyperref]{ranlp2021}
\usepackage{times}
\usepackage{latexsym}

\usepackage{microtype}
\usepackage{graphicx}
\usepackage{amsmath}
\usepackage{amsfonts}
\usepackage{booktabs}
\usepackage{multirow}

\let\ACMmaketitle=\maketitle
\renewcommand{\maketitle}{\begingroup\let\footnote=\thanks \ACMmaketitle\endgroup}

\title{Sentence Structure and Word Relationship Modeling \\ for Emphasis Selection \footnote{The work described in this paper is substantially supported by a grant from
the Direct Grant of the Faculty of Engineering, CUHK (Project Code: 4055093).}}

\author{Haoran Yang, Wai Lam\\
  The Chinese University of Hong Kong  \\
  \texttt{\{hryang, wlam\} @se.cuhk.edu.hk} \\
 }

\date{}

\begin{document}
\maketitle
\begin{abstract}
Emphasis Selection is a newly proposed task which focuses on choosing words for emphasis in short sentences. Traditional methods only consider the sequence information of a sentence while ignoring the rich sentence structure and word relationship information.  In this paper, we propose a new framework that considers sentence structure via a sentence structure graph and word relationship via a word similarity graph. The sentence structure graph is derived from the parse tree of a sentence. The word similarity graph allows nodes to share information with their neighbors since we argue that in emphasis selection, similar words are more likely to be emphasized together. Graph neural networks are employed to learn the representation of each node of these two graphs. Experimental results demonstrate that our framework can achieve superior performance. Code is available \url{https://github.com/LHRYANG/emphasis-selection}.
\end{abstract}

\section{Introduction}
Emphasis Selection recently proposed  by \cite{shirani-etal-2019-learning} aims to select candidate words for emphasis in short sentences. By emphasizing words, people's intent can be better conveyed, which is useful in a variety of applications. For example, it can be used in spoken language processing to generate more expressive sentences and be used to enable automated design  assistance  in  authoring, i.e., labeling important parts in a paragraph or in a poster title. 
Although it seems that this task is highly similar to the task of keyword extraction ~\cite{Beliga20141KE}, these two tasks are fundamentally different. The first difference is that keyword extraction focuses on a paragraph which is composed of multiple sentences while
emphasis selection aims to choose words from a short sentence. This difference implies that modeling sentence structure is more effective in emphasis selection. 
The second difference is that many global word statistics methods employed in keyword extraction such as  TF-IDF and word co-occurence frequency  will not work in this task, because for short sentences, it is  meaningless to count word frequency and whether the word should be emphasized has nothing to do with the frequency of the word. In addition, keyword extraction requires that the collected keywords are diverse, which means that if two words have similar meaning, only one should be kept. However, in emphasis selection, similar words tend to be emphasized together. Emphasis selection also shares some resemblances with entity recognition~\cite{yadav-bethard-2018-survey}. But one major difference is that 
the parts of speech of emphasized words are more diverse and the relation of adjacent words is weaker in the emphasis selection task.
\begin{figure}[!t]
 \includegraphics[width=\linewidth]{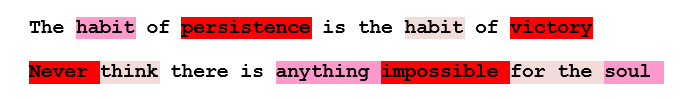}
  \caption{Two examples of the emphasis selection. Words with darker background indicate that more people agree to emphasize.}
  \label{fig:example}
  \vspace{-0.2cm}
\end{figure}

Generally speaking, emphasis selection can be modeled as a sequence classification task where the input is a sentence and the output is each word's probability to be emphasized. Shirani et al. ~\cite{shirani-etal-2019-learning} propose a model which is based on the Recurrent Neural Network~\cite{conf/interspeech/MikolovKBCK10} and KL-Divergence loss function. 
Despite the fact that it looks like a straightforward task, there still exist some challenges. 
The first challenge is about how to incorporate  sentence structure information into the model. Sentence structure information includes what role (subject, predicate, object, etc.) the word plays as well as  the position of the word in a sentence. Obviously, this kind of information is very useful. Existing works ~\cite{shirani-etal-2019-learning} fail to model the global structure of a sentence.
The second challenge is that there is no given context except a short sentence, so it requires the model to be able to capture some common patterns or regularities of most people.
More concretely, if two words are similar, they are more likely to be emphasized together. For example, in Figure~\ref{fig:example}, \textbf{persistence} and \textbf{victory} are more likely to be emphasized together. This observation can also be found in the second example: \textbf{Never} and \textbf{impossible}. Moreover, we analyze the training dataset and get a more concrete understanding of this phenomenon through the following procedures: For each training sentence, we consider the most popular emphasized word called word A. Then, we identify the most similar word called word B to the word A based on GloVe embedding ~\cite{pennington2014glove}. We find that the word B is also emphasized with a higher probability than other words in this sentence and this phenomenon occurs in about 26\% of the training dataset.
Therefore, modeling this kind of relationship between words definitely can help improve the performance of models. 

\begin{figure}[t]
  \centering
  \includegraphics[width=0.8\linewidth]{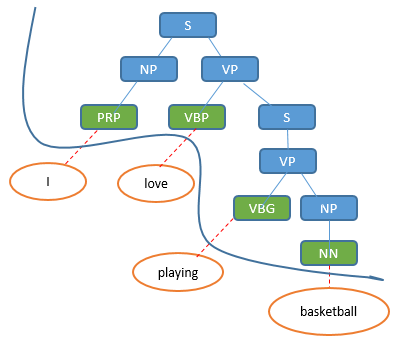}
  \caption{The part above the curve is the sentence structure graph constructed from the sentence: I love playing basketball. After the whole parse tree is encoded, the embeddings of the green nodes are used as the structure information for further classification.}
  \label{fig:Tag Graph}
  \vspace{-3pt}
\end{figure}

In this paper, we propose a sentence structure graph to handle the sentence structure issue.
Specifically, the sentence structure graph is derived from the parse tree of a sentence which contains useful information for this task. For example, as illustrated in Figure ~\ref{fig:Tag Graph}, when the path is S$\rightarrow$NP$\rightarrow$PRP, the word \textbf{I} is not inclined to be emphasized since this path indicates that this word is a subject. However, when the path is S$\rightarrow$VP$\rightarrow$S$\rightarrow$VP$\rightarrow$NP$\rightarrow$NN, the word \textbf{basketball} is likely to be emphasized since this word is a noun in a verb phrase. Generally, such sentence structure graph can reveal the role of words in a sentence which is beneficial for emphasis selection. Another important information - word relationship information is captured by a
 word similarity graph.
 Through the word similarity graph, words can share information with their neighbours, resulting in similar emphasized probabilities  of similar words. Next, graph neural networks~\cite{DBLP:journals/corr/VaswaniSPUJGKP17,cai-lam-2020-graph,DBLP:journals/corr/KipfW16,unknown,yun2019graph,vel2018graph} which has been demonstrated effective in modeling graph structure data are employed to learn the representation of each node of these two graphs. 
 We conduct extensive experiments based on different word embeddings, i.e., GloVe~\cite{pennington2014glove}, ELMo~\cite{peters2018contextualized}, RoBERTa~\cite{roberta} and the  experimental results show that our model can achieve superior performance.

\section{Related Work}
Emphasis selection is a new task proposed by ~\cite{shirani-etal-2019-learning} which aims to choose a subset of words to emphasize in a sentence. 
Shirani et al. ~\cite{shirani-etal-2019-learning} propose a model which is based on the Recurrent Neural Network~\cite{conf/interspeech/MikolovKBCK10}.
KL-Divergence loss function is adopted to conduct the label distribution learning (LDL)~\cite{DBLP:journals/corr/GengZ14}. This method achieves competitive performance over the sequence labeling model: CRF~\cite{Lafferty:2001:CRF:645530.655813}. 

In Recent years, graph neural networks ~\cite{unknown,DBLP:journals/corr/KipfW16,yun2019graph,vel2018graph,cai-lam-2020-graph} have demonstrated superiority in modeling the structure of graphs.  Kipf et al.~\cite{DBLP:journals/corr/KipfW16} propose a graph convolutional  network which is based on the fourier theory. One drawback of this model is that the edge weight of the graph needs to be known in advance. To overcome this shortcoming, Petar et al.~\cite{vel2018graph}  use a masked self-attention layer to calculate the weight of node's neighbours dynamically and then aggregate information by conducting a weighted addition operation. Currently, graph neural networks are applied to various tasks. Feria et al. ~\cite{DBLP:journals/corr/abs-1807-03012} construct a word graph by calculating the word embedding similarity and apply the community detection algorithm to find different communities. Through the graph, they can find named  entities  for  a bilingual language base in an unsupervised manner. Sun et al.~\cite{DBLP:journals/corr/abs-1905-07689} put forward a diverse graph pointer network for keyword extraction. They first construct a word graph based on the distance of two words and then use the graph convolutional network as an encoder to obtain each node's representation, finally a pointer network decoder and the diverse mechanism are employed to generate diverse keywords. The graph encoder can capture  document-level word salience and overcome the long-range dependency problem of RNN.

\section{Methodology}
We follow the same problem setting given by ~\cite{shirani-etal-2019-learning}. Suppose a sentence is composed of $n$ words $C=(x_1,x_2,...,x_n)$.
Our goal is to obtain a subset $S$ of words  in $C$ as selected words for emphasis where $1\leq \left|S\right| \leq n$.

We model this task as a prediction problem:
\begin{equation}
    (p_1,p_2,..,p_n) = model(x_1,x_2,...,x_n)
\end{equation}
where $p_i$ is $i$-th word's probability to be emphasized. Then $S$ contains the top-$\left|S\right|$ words with high probability.

\begin{figure}
  \centering
  \includegraphics[width=\linewidth]{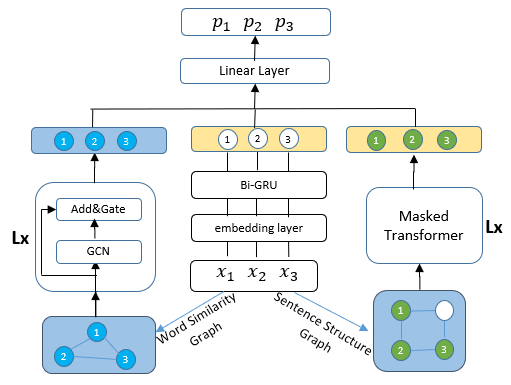}
  \caption{An overview of our model, the left part is the word similarity graph encoder, the middle part is the sequence encoder and the right part is the sentence structure graph encoder. Lx representes that there are L such blocks.}
  \label{fig:model}
\end{figure}

Figure~\ref{fig:model} depicts the architecture of our proposed model which is composed of three parts: (i) the middle part - sequence encoder (ii) the left part - word similarity graph encoder (iii) the right part - sentence structure graph encoder. Next, we will provide a detailed description of each part.
\subsection{Sequence Encoder}
The sequence encoder is composed of an embedding layer and a bidirectional GRU. It is mainly used to model the sequence information, i.e., word sequence and tag sequence. Formally, 
given a sentence $C=(x_1,x_2,...,x_n)$ with $n$ words, the embedding layer is responsible for converting each word into a $d_1$-dimensional vector and converting the corresponding POS tag into a $d_2$-dimensional vector:
\begin{equation}
    (w_1,...,w_n) = WordEmbed(x_1,...,x_n) 
\end{equation}
\begin{equation}
    (e_1,..,e_n) = TagEmbed(t_1,...,t_n)
\end{equation}
where $(t_1,...,t_n)$ is the POS tag sequence and $w_i \in \mathbb{R}^{d_1}, e_i \in  \mathbb{R}^{d_2}$. Then the word embedding and the tag embedding are concatenated and fed into a encoder $E$ to encode the sequence information. We can obtain the outputted hidden state of the encoder:
\begin{equation}
    (h_1,...,h_n) = E([w_1,e_1],...[w_n,e_n])
\end{equation}
\subsection{Word Relationship Modeling}
Given a sentence, we take each word as a node and the weight of the edge is calculated by the word embedding similarity. The weight matrix is denoted by $A \in \mathbb{R}^{n\times n}$. 
After the graph is constructed, a $L$-layer graph convolutional  network (GCN) ~\cite{DBLP:journals/corr/KipfW16} is employed to encode the word similarity graph:
\begin{equation}
    H^{l+1} = ReLU(D^{-\frac{1}{2}}AD^{-\frac{1}{2}}H^lW^l \label{GCN})
\end{equation}
where $W^l$ is a parameter and $H^l$ denotes the nodes' representation in the $l$-th layer. $D \in \mathbb{R}^{n\times n}$ is a diagonal matrix and $D_{ii}=\sum_j A_{ij}$. 

 Recall that the WSG is a complete graph since each two words are connected by a weighted edge. There exists a serious problem: Useful information may be overwhelmed by useless information, because a majority number of words do not need to be emphasized, causing the information in words that are not emphasized dominates the words that should be emphasized. To alleviate this problem, we adopt two strategies:  residual module ~\cite{DBLP:journals/corr/HeZRS15} and gate mechanism~\cite{DBLP:journals/corr/GehringAGYD17,DBLP:journals/corr/DauphinFAG16}. The residual module makes the current node's representation as the addition between the former representation and the aggregated information from its neighbours. The gate mechanism controls the magnitude of the aggregated information. Through this way, the current node's representation will not be significantly affected by its neighbours.
 Therefore Equation \eqref{GCN} can be rewritten as:
 \begin{align}
& \qquad \qquad M^{l+1} = H^lW^l \\
& \qquad \qquad C = s(H^lW^g) \\
&H^{l+1} = M^{l+1} +D^{-\frac{1}{2}}AD^{-\frac{1}{2}}M^{l+1}\otimes C 
 \end{align}
where  $s(\cdot)$ is the sigmoid function and $\otimes$ is the point-wise multiplication.

We obtain $H^0$ from the word embedding matrix and obtain the $L$-th layer output $H^L=(w^L_1,...,w^L_n)$ as each node's features of the word similarity graph.

\subsection{Sentence Structure Modeling}
SSG is constructed by parsing the sentence using NLTK\footnote{\url{https://www.nltk.org/}} and StandfordNLP~\footnote{\url{https://stanfordnlp.github.io/CoreNLP/}}. Then, we remove the leaf nodes (which are the words) and  the remaining part is the SSG. Each node of the graph is a kind of POS tag and the path from the root to a specific word can reveal what role the word plays in the sentence.  

 Apparently, the weight of edges is important. For example, in Figure~\ref{fig:Tag Graph}, the root node S has two children nodes NP and VP. The edge (S, NP) should have a smaller weight than the edge (S, VP) since people tend not to emphasize the subject in most circumstances. Different from WSG where the weight can be calculated by the word embedding similarity explicitly, it is not appropriate to calculate the weight in the SSG by the node similarity.
Hence, we integrate the idea of Transformer~\cite{DBLP:journals/corr/VaswaniSPUJGKP17,cai-lam-2020-graph} and masked self-attention ~\cite{vel2018graph} to the SSG modeling.
Firstly, we generate three vectors: key, query, value, according to the current node's representation:

\begin{align}
k^{l+1}_{i} = \mathbf{W}^l_k(v^l_{i}) \\ 
q^{l+1}_{i} = \mathbf{W}^l_q(v^l_{i})   \\
 v^{l+1}_{i} = \mathbf{W}^l_v(v^l_{i}) 
\end{align}
where $\mathbf{W}^l_k, \mathbf{W}^l_q,  \mathbf{W}^l_v$ are parameters. $k^{l}_{i},q^{l}_{i},v^{l}_{i}$ correspond to the $l$-th layer key, query, value vector respectively. $v^0_i$ is initialized  from the tag embedding matrix. Then, a masked self-attention is employed to allow nodes aggregating information only from their neighbours.

\begin{equation}
    v^{l+1}_i = \sum_{j \in \mathcal{N}(i)} a_{ij}v^{l+1}_j
\end{equation}
\begin{equation}
    a_{ij} = \frac{exp(q^{l+1}_i k^{l+1}_j)}{\sum_{z \in \mathcal{N}(i)} exp(q^{l+1}_i k^{l+1}_z)}
\end{equation}
where $ \mathcal{N}(i)$ is the neighbour set of the node $i$.  After the graph is encoded with a $L$-layer network, we obtain the leaf nodes (the green nodes shown in Figure ~\ref{fig:Tag Graph})  representation  $V=(v^L_1,v^L_2,...,v^L_n)$.

\subsection{Loss Function}
After obtaining these three modules' output, we conduct a concatenation operation and calculate the probability:

\begin{equation}
    p_i = softmax ( f([h_i,v^L_i,w^L_i]) )
\end{equation}
where $p_i \in \mathbb{R}^3$ is $i$-th word's probability distribution. $f$ represents a fully connected  neural network.
We adopt negative log likelihood as the loss function:

\begin{equation}
    L = -\sum_{C \in D_{train}}\sum_{i=1}^{\left|C\right|} \log p_{iy_i}
\end{equation}
\section{Experiment and Results}
\begin{table}[h]
\small
\centering
\begin{tabular}{lllllll}
\hline
\toprule
DIY &ideas&for&leafing&up&your&home\\
\hline
\midrule
B& O & O & O & O & B & I \\
B& I & O & O & O & O& O \\ 
B& O & O & O & O & O & O\\ 
O& B & O & O & O & O & O\\ 
O& O & O & B & O & O & B\\
O& O & O & B & I & I & I\\
O& O & O & B & O & O & O\\
B& O & O & B & O & O & B\\
B& I & O & O & O & O & O\\

\bottomrule
\hline 
\end{tabular}
\caption{An example of the labeled dataset}
\label{tab:one_example}
\end{table}

\begin{table*}[t]
\centering

\begin{tabular}{llllll}
\hline 
\toprule
Methods &Match-1&Match-2&Match-3&Match-4&Average\\
\hline 
GloVe \\
\hline
CNN & 0.541 & 0.678 & 0.754 & 0.805 & 0.695 \\ 
RNN~\cite{shirani-etal-2019-learning}       & 0.536 & \textbf{0.712} & 0.777 & 0.811 & 0.709 \\
Ours     &\textbf{0.569} & 0.703 & 0.772 & \textbf{0.813} & 0.714  \\
Ours w/o WSG     & 0.563 & 0.710 & \textbf{0.778} & 0.810 & \textbf{0.715} \\
Ours w/o SSG     & 0.561 & 0.710 & 0.769 & 0.811 & 0.713\\
\hline 
\midrule
ELMo \\
\hline 
CNN & 0.574 & 0.729 & 0.795 & 0.832 & 0.733\\
RNN-based~\cite{shirani-etal-2019-learning} & 0.592 & 0.752 & 0.804 & 0.822 & 0.743 \\
Ours     & \textbf{0.610} & \textbf{0.768} & \textbf{0.813} & \textbf{0.836} & \textbf{0.757} \\
Ours w/o WSG         & 0.604 & 0.742 & 0.804 & 0.827 & 0.744 \\
Ours w/o SSG       & 0.597 & 0.753 & 0.801 & 0.836 & 0.747 \\ 

\bottomrule
\hline 
\end{tabular}
\caption{Results of our model and baselines on GloVe and ELMo. The best performance is boldfaced.}
\label{tab:results}
\end{table*}

\begin{table*}[t]
\centering
\begin{tabular}{c|cccccc}
\hline 
\toprule
&Stay &foolish&to&stay&sane&.\\
\hline 
\midrule
Annotator &0.333(4) &0.889(1/2)&0.222(5/6)&0.444(3)&0.889(1/2)&0.222(5/6)\\ 
RNN-based &0.502(3)&0.565(2)&0.227(6)&0.460(4)&0.798(1)&0.357(5) \\ 
Ours & 0.502(4) &0.784(2) &0.210(6)&0.595(3)&0.805(1)&0.288(5)\\ 

\bottomrule
\hline 
\end{tabular}
\caption{A sample case. 
Numbers outside the brackets indicate the word's probability of being emphasized. Numbers in the brackets are the ranking of the corresponding word. (a/b) means that two words have the same ranking.}  
\label{tab:case}
\end{table*}

\subsection{Dataset}
We use the dataset\footnote{\url{https://github.com/RiTUAL-UH/SemEval2020_Task10_Emphasis_Selection}} provided by ~\cite{shirani-etal-2019-learning}. The dataset contains $2742$ training sentences and 392  test sentences. Each sentence is labeled by nine annotators. Table ~\ref{tab:one_example} gives a sample record of one sentence. B, I, O represent the beginning word to be emphasized, the interior word to be emphasized, and the word not to be emphasized respectively. Since there exists different opinions about whether the word should be emphasized, the labels given by nine annotators are slightly different.

\subsection{Experimental Setup}
 We regard each annotator's labeling as a sample in the dataset. In other words, each sentence is associate with nine samples. 
In order to verify the robustness of our model, we conduct experiments on two pre-trained word embeddings: 300-$d$ GloVe~\cite{pennington2014glove} and 2048-$d$ ELMo~\cite{peters2018contextualized}. For the above two kinds of embeddings, we adopt GRU as the encoder $E$. The GRU hidden state size is $512$ and $1024$ respectively. The word similarity graph's node embedding size is $300$ and $2048$ respectively. The sentence structure graph's node embedding size is $300$ and $512$ respectively. Moreover, we  initialize the sentence structure graph's node embedding by training a  classifier which only uses the sentence structure graph encoder. We adopt a two-layer bidirectional GRU. The sentence structure graph and the word similarity graph are encoded by a two-layer graph neural network. The batch size is set to 16. The negative slope of the ReLU function is set to $0.2$. We use the Adam optimizer and the learning rate is $0.0001$. The number of epoch is $100$. We also add a dropout layer and the dropout rate is $0.5$. 

Since generalized pretrained language models such as BERT~\cite{bert}, RoBERTa~\cite{roberta} are demonstrated effective in a large bunch of  downstream tasks, we also report results obtained by fine-tuning the RoBERTa on the emphasis selection dataset. There are two different experimental settings. The first setting is that only the RoBERTa model is used as the encoder $E$. The second setting is that a GRU layer is added on the top of the RoBERTa model, i.e., RoBERTa+GRU is the encoder $E$. The sentence structure model and the word relationship model remain unchanged. Adam optimizer is adopted and the learning rate is set to 1e-5. 
\subsection{Evaluation Metric}
We adopt \textbf{Match-m}  ~\cite{shirani-etal-2019-learning} as the evaluation metric which is defined as below:
\begin{table*}[t]
\centering

\begin{tabular}{llllll}
\hline 
\toprule
Methods &Match-1&Match-2&Match-3&Match-4&Average\\
\hline 
RoBERTa \\
\hline
Ours w/o both     &0.635 & 0.756 & 0.803 & 0.832 & 0.757  \\
Ours w/o WSG     & \textbf{0.640} & 0.775 & 0.793 & 0.827  & 0.759 \\
Ours w/o SSG     & 0.633 & 0.760 & \textbf{0.804} & \textbf{0.839} & 0.759\\
Ours    &0.633 &  \textbf{0.779} & 0.803 & 0.833 & \textbf{0.762}\\
\hline 
\midrule
RoBERTa+GRU \\
\hline 
Ours w/o both    & 0.607 & 0.755 & 0.795 & 0.822 & 0.745 \\
Ours w/o WSG         &  0.602 & \textbf{0.766} & 0.798 & 0.825 & 0.748 \\
Ours w/o SSG       & \textbf{0.607} & 0.758 & 0.801 & 0.837 & 0.747 \\ 
Ours & 0.600 & 0.761 & \textbf{0.806} & \textbf{0.838} & \textbf{0.751}\\
\bottomrule
\hline 
\end{tabular}
\caption{Results of our model and baselines based on two different architectures, RoBERTa and RoBERTa+GRU. The best performance is boldfaced.}
\label{tab:results_lm}
\end{table*}

\begin{table*}[t]
\centering
 \scalebox{0.9}{\begin{tabular}{c|ccccccccc}
\hline 
\toprule
&Thanks &for &showing &me &all &the & best &dance &moves\\
\hline 
\midrule
Annotator &0.444 &0.111&0.111&0.111&0.111&0&0.888&0.555&0.444\\ 
Ours & 0.412 &0.057 &0.332&0.092&0.063&0.024&0.406&0.599&0.387\\ 

\bottomrule
\hline 
\end{tabular}}
\caption{A failed case. 
Numbers are the word's probability of being emphasized.}  
\label{tab:case_failed}
\end{table*}
 \textbf{Match-m}: For a sentence $C$, 
 we choose $m$ words (denoted by $S^g_m(C)$) with the top-$m$ probability (probability of the label B + probability of the label I) in the ground truth and $m$ words (denoted by $S^p_m(C)$)  based on the predicted probability. The formula is defined as:
\begin{equation}
\text{Match-m} = \frac{\sum_{C \in D_{test}} \frac{\left| S^g_m(C) \cap S^p_m(C)\right|}{min(\left|C\right|,m)}}{\left|D_{test} \right|}
\end{equation}

\subsection{Results and Analysis}
\subsubsection{Experimental Results}
We compare our model with the existing model based on RNN proposed by ~\citet{shirani-etal-2019-learning} and the convolutional neural network (CNN). We report results evaluated by the metrics Match-1, Match-2, Match-3, Match-4 and the average of these four metrics.
From Table \ref{tab:results}, we can see that CNN lags behind  other models on the whole.  

When the word embedding is GloVe, models with at least one graph surpass RNN on almost all the metrics except Match-2. In particular, our model can achieve an improvement on Match-1 and Match-4. Our model without WSG (word similarity graph) achieves an excellent performance on Match-3 and Average. When the word embedding is ELMo, 
ours is superior to RNN-based on all the evaluation metrics. Compared to these two ablated models, Ours can also achieve better performance. Ours w/o WSG is better than RNN-based on all the evaluation metrics except Match-2 and Ours w/o SSG is better than RNN-based except Match-3.   
On the whole, models with graphs can obtain better results on most metrics compared to the baseline models, which shows the advantage of these two components. 

Experimental results based on RoBERTa are listed in Table~\ref{tab:results_lm}. Compared with the results based on GloVe and ELMo, RoBERTa and its variants achieve higher average match score which shows that a better initialized word embedding is helpful for a better performance.  For the same RoBERTa encoder, Ours can obtain the highest score on Average and Match-2. For RoBERTa+GRU encoder, Ours can obtain the highest score on Average, Match-3 and Match-4. However, one interesting finding is that RoBERTa encoder performs much better than RoBERTa+GRU encoder. Two possible reasons may interpret this phenomenon. The first reason is the overfitting problem and the second reason is that the larger network is  harder to train due to some optimization issues, e.g., gradient vanishing.  
\subsubsection{Case Study} 
To gain some insights of our proposed model, we present a sample case generated by the ELMo-based model as shown in Table~\ref{tab:case}. We can see that Ours not only predicts the ranking accurately, but also obtains very close probability to the ground truth probability derived by annotators. 
Besides that, the probabilities of foolish and sane predicted by our model are very close than that predicted by RNN-based, which shows that the word similarity graph can impel similar words to have similar probabilities. 

We also provide a failed case in Table~\ref{tab:case_failed}. It is intrinsically harder to rank the words in this sentence even for human beings. Our model does not rank them correctly on these cases where multiple words may be emphasized.   
\subsubsection{Some Useful Tips}
We conclude some tips on the experiment that leads to better performance. (1) We can firstly train a classifier only using the SSG, then use the pre-trained embeddings as an initialization  of the sentence graph nodes embeddings. It can obtain higher score and faster convergence of the model. 
(2) We also consider another method to model the relationships between words using a self-attention operation proposed by ~\citet{article} above the hidden vectors of RNN. However, the performance is slightly degraded compared to removing this operation. So we think it is much better to model words relationships and sequence information separately.     
\section{Conclusions}
The sentence structure graph and the word similarity graph are proposed to solve two issues found in emphasis selection. 
The sentence structure graph helps to model the structure information of the sentence and the word similarity graph is useful in
modeling relationships between words.
With the development of graph neural network, the two graphs can be properly encoded and integrated into existing models.  Experimental results demonstrate that our framework can achieve superior performance. 
\bibliographystyle{acl_natbib}
\bibliography{ranlp2021}

\begin{thebibliography}{22}
\expandafter\ifx\csname natexlab\endcsname\relax\def\natexlab#1{#1}\fi

\bibitem[{Cai and Lam(2020)}]{cai-lam-2020-graph}
Deng Cai and Wai Lam. 2020.
\newblock Graph transformer for graph-to-sequence learning.
\newblock In \emph{34th AAAI Conference on Artifical Intelligence}.

\bibitem[{Dauphin et~al.(2017)Dauphin, Fan, Auli, and
  Grangier}]{DBLP:journals/corr/DauphinFAG16}
Yann~N. Dauphin, Angela Fan, Michael Auli, and David Grangier. 2017.
\newblock Language modeling with gated convolutional networks.
\newblock In \emph{34th International Conference on Machine Learning (ICML)},
  page 933–941.

\bibitem[{Devlin et~al.(2019)Devlin, Chang, Lee, and Toutanova}]{bert}
Jacob Devlin, Ming-Wei Chang, Kenton Lee, and Kristina Toutanova. 2019.
\newblock \href {https://doi.org/10.18653/v1/N19-1423} {{BERT}: Pre-training of
  deep bidirectional transformers for language understanding}.
\newblock In \emph{Proceedings of the 2019 Conference of the North {A}merican
  Chapter of the Association for Computational Linguistics: Human Language
  Technologies, Volume 1 (Long and Short Papers)}, pages 4171--4186,
  Minneapolis, Minnesota. Association for Computational Linguistics.

\bibitem[{Feria et~al.(2018)Feria, Balbin, and
  Bautista}]{DBLP:journals/corr/abs-1807-03012}
Miguel Feria, Juan~Paolo Balbin, and Francis~Michael Bautista. 2018.
\newblock Constructing a word similarity graph from vector based word
  representation for named entity recognition.

\bibitem[{Gehring et~al.(2017)Gehring, Auli, Grangier, Yarats, and
  Dauphin}]{DBLP:journals/corr/GehringAGYD17}
Jonas Gehring, Michael Auli, David Grangier, Denis Yarats, and Yann~N. Dauphin.
  2017.
\newblock Convolutional sequence to sequence learning.
\newblock In \emph{34th International Conference on Machine Learning (ICML)},
  page 1243–1252.

\bibitem[{Geng and Zhao(2014)}]{DBLP:journals/corr/GengZ14}
Xin Geng and Quan Zhao. 2014.
\newblock Label distribution learning.

\bibitem[{Gupta(2017)}]{Beliga20141KE}
Er.~Tanya Gupta. 2017.
\newblock Keyword extraction: A review.
\newblock In \emph{IJEAST}, pages 215--220.

\bibitem[{He et~al.(2016)He, Zhang, Ren, and Sun}]{DBLP:journals/corr/HeZRS15}
Kaiming He, Xiangyu Zhang, Shaoqing Ren, and Jian Sun. 2016.
\newblock Deep residual learning for image recognition.
\newblock In \emph{2016 IEEE Conference on Computer Vision and Pattern
  Recognition (CVPR)}, pages 770--778.

\bibitem[{Kipf and Welling(2017)}]{DBLP:journals/corr/KipfW16}
Thomas~N. Kipf and Max Welling. 2017.
\newblock Semi-supervised classification with graph convolutional networks.
\newblock In \emph{5th International Conference on Learning Representations
  (ICLR)}.

\bibitem[{Lafferty et~al.(2001)Lafferty, McCallum, and
  Pereira}]{Lafferty:2001:CRF:645530.655813}
John~D. Lafferty, Andrew McCallum, and Fernando C.~N. Pereira. 2001.
\newblock Conditional random fields: Probabilistic models for segmenting and
  labeling sequence data.
\newblock In \emph{ICML}, pages 282--289.

\bibitem[{Lin et~al.(2017)Lin, Feng, Dos~Santos, Yu, Xiang, Zhou, and
  Bengio}]{article}
Zhouhan Lin, Minwei Feng, Cicero Dos~Santos, Mo~Yu, Bing Xiang, Bowen Zhou, and
  Y.~Bengio. 2017.
\newblock A structured self-attentive sentence embedding.
\newblock In \emph{5th International Conference on Learning Representations
  (ICLR)}.

\bibitem[{Liu et~al.(2019)Liu, Ott, Goyal, Du, Joshi, Chen, Levy, Lewis,
  Zettlemoyer, and Stoyanov}]{roberta}
Yinhan Liu, Myle Ott, Naman Goyal, Jingfei Du, Mandar Joshi, Danqi Chen, Omer
  Levy, Mike Lewis, Luke Zettlemoyer, and Veselin Stoyanov. 2019.
\newblock \href {http://arxiv.org/abs/1907.11692} {Roberta: {A} robustly
  optimized {BERT} pretraining approach}.
\newblock \emph{CoRR}, abs/1907.11692.

\bibitem[{Mikolov et~al.(2010)Mikolov, Karafiát, Burget, Cernocký, and
  Khudanpur}]{conf/interspeech/MikolovKBCK10}
Tomas Mikolov, Martin Karafiát, Lukás Burget, Jan Cernocký, and Sanjeev
  Khudanpur. 2010.
\newblock Recurrent neural network based language model.
\newblock In \emph{INTERSPEECH}, pages 1045--1048.

\bibitem[{Pennington et~al.(2014)Pennington, Socher, and
  Manning}]{pennington2014glove}
Jeffrey Pennington, Richard Socher, and Christopher~D Manning. 2014.
\newblock Glove: Global vectors for word representation.
\newblock In \emph{EMNLP}, pages 1532--1543.

\bibitem[{Peters et~al.(2018)Peters, Neumann, Iyyer, Gardner, Clark, Lee, and
  Zettlemoyer}]{peters2018contextualized}
Matthew~E. Peters, Mark Neumann, Mohit Iyyer, Matt Gardner, Christopher Clark,
  Kenton Lee, and Luke Zettlemoyer. 2018.
\newblock Deep contextualized word representations.
\newblock In \emph{NAACL}, page 2227–2237.

\bibitem[{Shirani et~al.(2019)Shirani, Dernoncourt, Asente, Lipka, Kim,
  Echevarria, and Solorio}]{shirani-etal-2019-learning}
Amirreza Shirani, Franck Dernoncourt, Paul Asente, Nedim Lipka, Seokhwan Kim,
  Jose Echevarria, and Thamar Solorio. 2019.
\newblock Learning emphasis selection for written text in visual media from
  crowd-sourced label distributions.
\newblock In \emph{Proceedings of the 57th Annual Meeting of the Association
  for Computational Linguistics}, page 1167–1172.

\bibitem[{Sun et~al.(2019)Sun, Tang, Du, Deng, and
  Nie}]{DBLP:journals/corr/abs-1905-07689}
Zhiqing Sun, Jian Tang, Pan Du, Zhi{-}Hong Deng, and Jian{-}Yun Nie. 2019.
\newblock Divgraphpointer: {A} graph pointer network for extracting diverse
  keyphrases.
\newblock In \emph{SIGIR}, pages 755--764.

\bibitem[{Vaswani et~al.(2017)Vaswani, Shazeer, Parmar, Uszkoreit, Jones,
  Gomez, Kaiser, and Polosukhin}]{DBLP:journals/corr/VaswaniSPUJGKP17}
Ashish Vaswani, Noam Shazeer, Niki Parmar, Jakob Uszkoreit, Llion Jones,
  Aidan~N. Gomez, Lukasz Kaiser, and Illia Polosukhin. 2017.
\newblock Attention is all you need.
\newblock In \emph{NIPS}, pages 5998--6008.

\bibitem[{Veličković et~al.(2018)Veličković, Cucurull, Casanova, Romero,
  Liò, and Bengio}]{vel2018graph}
Petar Veličković, Guillem Cucurull, Arantxa Casanova, Adriana Romero, Pietro
  Liò, and Yoshua Bengio. 2018.
\newblock Graph attention networks.
\newblock In \emph{6th International Conference on Learning Representations
  (ICLR)}.

\bibitem[{Wu et~al.(2019)Wu, Pan, Chen, Long, Zhang, and Yu}]{unknown}
Zonghan Wu, Shirui Pan, Fengwen Chen, Guodong Long, Chengqi Zhang, and
  Philip~S. Yu. 2019.
\newblock \href {http://arxiv.org/abs/1901.00596} {A comprehensive survey on
  graph neural networks}.
\newblock \emph{CoRR}, abs/1901.00596.

\bibitem[{Yadav and Bethard(2018)}]{yadav-bethard-2018-survey}
Vikas Yadav and Steven Bethard. 2018.
\newblock A survey on recent advances in named entity recognition from deep
  learning models.
\newblock In \emph{COLING}, pages 2145--2158.

\bibitem[{Yun et~al.(2019)Yun, Jeong, Kim, Kang, and Kim}]{yun2019graph}
Seongjun Yun, Minbyul Jeong, Raehyun Kim, Jaewoo Kang, and Hyunwoo~J. Kim.
  2019.
\newblock Graph transformer networks.
\newblock In \emph{NIPS}, pages 11983--11993.

\end{thebibliography}

\end{document}